# Multi-focus Attention Network for Efficient Deep Reinforcement Learning

**Jinyoung Choi[1], Beom-Jin Lee[2], and Byoung-Tak Zhang[2]**

[1]Interdisciplinary Program in Cognitive Science, Seoul National University
[2]School of Computer Science and Engineering, Seoul National University
{jychoi,bjlee,btzhang}@bi.snu.ac.kr

## Abstract

Deep reinforcement learning (DRL) has shown incredible performance in learning various tasks to the human level. However, unlike human perception, current DRL models connect the entire low-level sensory input to the state-action values rather than exploiting the relationship between and among entities that constitute the sensory input. Because of this difference, DRL needs vast amount of experience samples to learn. In this paper, we propose a Multi-focus Attention Network (MANet) which mimics human ability to spatially abstract the low-level sensory input into multiple entities and attend to them simultaneously. The proposed method first divides the low-level input into several segments which we refer to as partial states. After this segmentation, parallel attention layers attend to the partial states relevant to solving the task. Our model estimates state-action values using these attended partial states. In our experiments, MA-Net attains highest scores with significantly less experience samples. Additionally, the model shows higher performance compared to the Deep Q-network and the single attention model as benchmarks. Furthermore, we extend our model to attentive communication model for performing multi-agent cooperative tasks. In multi-agent cooperative task experiments, our model shows 20% faster learning than existing state-of-the-art model.

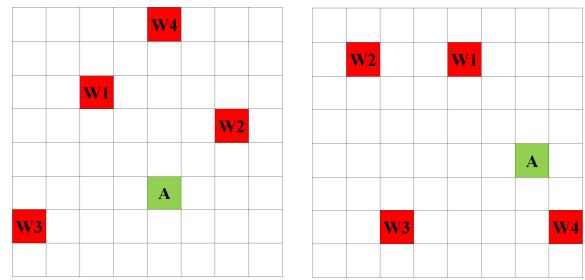

*Figure 1. Examples of grid world navigation. In spite of same relative position of A and W1, DQN has to learn these cases independently.*

## Introduction

Deep reinforcement learning has shown incredible performance in learning various tasks to the human level. Deep Q-network (DQN) successfully learned to play Atari 2600 games and achieved human-level scores in many games (Mnih et al. 2015) and Zhang et al. (2015) used DQN for robotic arm control attaining high performance results. However, DQN connects entire low-level sensory input to the state-action values rather than exploiting the relationship inter and intra entities that constitute the sensory input.

For example, assume the grid world navigation task using the map (Figure 1). The agent receives the image representing the whole grid as input and each cell in the grid is rendered as a small box in the image. The agent (A) visits waypoints (W1, W2, W3, W4) in an orderly sequence. In two images (Figure 1), the positions of A and W1 are different but their relative positions are same. Therefore, if a RL model can attend to A and W1, it may learn the task faster by applying same policy. However, DQN perceives two images in Figure 1 as completely different states. Therefore, DQN has to learn policy about every possible placement of entities (A and Ws) independently. The number of possible placement of entities grows exponentially with the number of entities. Thus, DQN requires enormous experience samples to learn and training DQN is often intractable in real world environment where the agent cannot expect to experience exactly the same scene repetitively.

Recently, to overcome this limitation, works like (Sorokin et al. 2015; Oh et al. 2016) applied attention concepts to the DRL models in order to learn Atari 2600 and Minecraft maze task, respectively. However, their attention mechanisms used only one softmax layer. Hence, their models show significant difficulty in attending to multiple entities (or memories) with equal importance whereas humans

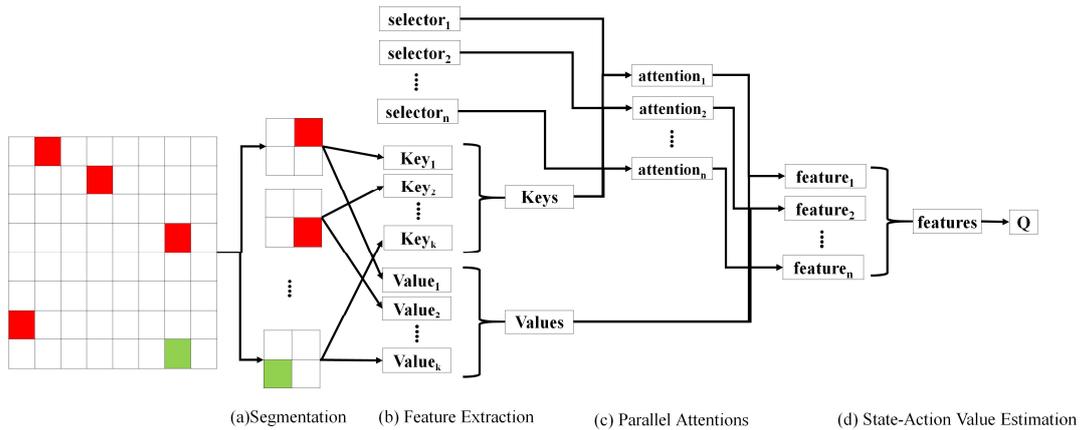

*Figure 2. Illustration of our model's structure.*

have ability to attend to multiple entities simultaneously (Finke et al. 2007).

In this paper, we propose a Multi-focus Attention Network (MANet) that enhances the agent's ability to attend to important entities by using multiple parallel attentions. Our model segments the low-level sensory input into multiple segments which we refer to as partial states. Parallel attention layers attend to partial states that are relevant to solving the task. Then the attended partial states estimate state-action values. In experiments, our model shows remarkably faster learning speed and higher performance than DQN and single attention model.

Furthermore, by viewing partial states as partially observed states perceived by multiple agents, we extend our model to the multi-agent reinforcement learning model. In the multi-agent cooperation task, our model learns roughly 20% faster than the state-of-the-art model (Sukhbaatar, Szlam, and Fergus 2016).

## Related Works

**Deep Reinforcement Learning:** Deep neural networks have been successful in supervised learning tasks like image classification (Krizhevsky, Sutskever, Hinton. 2012; He et al. 2015). Recently, (Mnih et al. 2015) proposed a Deep Q-Network(DQN), which used deep convolutional neural network for state-action value function approximation. DQN learned many Atari 2600 games to the human-level in an end-to-end fashion. Zhang et al. (2015) applied DQN to the robotic arm control and trained the robot arm using simulation images. However training DQN requires long time and enormous experiences, making the direct application to the real-world environment intractable. Many works have followed to accelerate learning speed of DQN. Mnih et al. (2016) proposed asynchronous learning methods and Nair et al. (2015) proposed massively parallel methods. Both methods boosted learning speed of DQN greatly. However, these methods focused on exploiting massive computing powers rather than improving the model's efficiency.

**Cognitive Science:** In developmental psychology literatures, there is evidence that human cognitive system is found on some core knowledge: entities, other agents (actions), numbers, spaces (Spelke and Kinzler 2007). Attention is considered important in perceiving coherent entities or agents. Furthermore, there is evidence that humans have the ability to attend to multiple entities simultaneously (Finke et al. 2007). Our work is inspired by these insights, focusing on simultaneous attention for fast and efficient learning.

**Attention in Deep Reinforcement Learning:** Recently, visual attention has been successful in many supervised learning tasks like caption generation (Xu et al. 2015), image generation (Gregor et al. 2015), language translation (Bahdanau et al. 2014). Following these achievements in supervised learning domain, Sorokin et al. (2015) proposed Deep Attention Recurrent Q-Network (DARQN), which integrated attention mechanism into DQN. DARQN surpassed the performance of DQN in some Atari 2600 games. However, it failed to surpass DQN in many Atari 2600 games. Oh et al. (2016) proposed Memory Q-Network (MQN), which used attention for context dependent memory retrieval. MQN outperformed DQN by a large margin in maze tasks, which require ability to memorize important information for long time-steps. In contrast to these works, we use attention to extract important partial states in the input of one time-step and apply multiple attention layers to attend to multiple partial states for fast and efficient learning

**Multi-Agent Communication:** In many multi-agent reinforcement learning models, communication between agents is usually pre-determined (Busoniu, Babuska, and De Schutter 2008). Recently, Sukhbaatar, Szlam, and Fergus (2016) proposed Communication Neural Net (CommNN). CommNN learns not only the policy for individual agents, but also the communication protocol between agents in an end-to-end fashion. CommNN outperformed non-communicative models and baselines in various multi-agent cooperative tasks. In contrast to CommNN, our model encodes two different feature. One estimates the state-action values and the other is used for communication between agents. Furthermore, agents attend to other agents that have important information. These attributes enable clear and rich communication between agents, leading to faster learning.

## Multi-focus Attention Network

**Background: Deep Q Learning**
We use Deep Q-Learning framework introduced by Mnih et al. (2015). Deep Q-Learning uses neural network to approximate the state-action value function. Let $s_t$ be a low-level sensory input at time step $t$, $a_t$ be a discrete action at $t$, $r_t$ be a reward at $t$. Then the loss function of Deep Q-Learning framework is defined as follows:

$$L = \{r_t + \gamma * \max_{a'} Q'(s_{t+1}, a') - Q(s_t, a_t)\}^2 \quad (1)$$

where $\gamma$ is the discount coefficient and $a'$ is the possible actions can be chosen at $t+1$, $Q$ is neural network (Q network) and $Q'$ is target network. The target network is used to stabilize learning. It has same structure to the Q network and weights of the Q network are copied to the target network periodically. We also use replay memory mechanism (Lin. 1993) which stores transitions $\{(s_t, a_t, r_t, s_{t+1})\}$ and uses randomly sampled transitions to train the network.

**Single Agent Setting**
Our model (Figure 2) consists of 4 modules: (a) Input segmentation, (b) Feature extraction, (c) Parallel attentions, (d) State-action value estimation.

**Input Segmentation:** Input segmentation module segments the low-level sensory input into multiple segments which we refer to as partial states. This can be done by various methods. In our experiments, we used simple grid method. We partitioned input image into uniform grid and used the cells (small image patches) in the grid as partial states. We believe that we can apply more sophisticated methods like super-pixel segmentation (Achanta et al. 2010) or spatial transformer networks (Jaderberg, Simonyan, and Zisserman 2015). We will explore the feasibility in future works.

**Feature Extraction:** The feature extraction module extracts key and value features from each of partial states. The key features are used to determine where the model should attend and the value features are used to encode information for state-action value estimation. This two stream feature extraction is inspired by (Oh et al. 2016). For each partial state, we first define common feature as follows:

$$c_i = f_f(s_i) \quad for\ all\ i \in (0,1,\ldots,K) \quad (2)$$

where K is the number of partial states, $c_i$ is the common feature for i-th partial state, $f_f$ is extraction function, and $s_i$ is i-th partial state. In our experiments, we used deep convolutional neural network and deep neural network as $f_f$. Additionally, we concatenated index of the partial state to the common feature. Using the extracted $c_i$, we define key and value feature as follows:

$$Key_i = W_{key} * c_i \quad (3)$$

$$Val_i = f_v(W_{val} * c_i) \quad (4)$$

where $Key_i$ is the key feature for i-th partial state, $Val_i$ is the value feature for i-th partial state, $f_v$ is non-linear activation function, and $W_{key}$, $W_{val}$ are the weight matrixes. We use leaky ReLU (Maas 2013) as $f_v$.

**Parallel Attentions:** Using the key features extracted from the feature extraction module, parallel attention layers determine what partial states are important by using the following equations:

$$A_i^n = \frac{\exp(a_n * Key_i^T)}{\sum_{i'} \exp(a_n * Key_{i'}^T)} \quad for\ all\ n \in (0,1,\ldots,N) \quad (5)$$

where $N$ is the number of attention layers, $A_i^n$ is i-th element of n-th soft attention weight vector, $i' \in \{0,1,\ldots,K\} - i$ and $a_n$ is n-th selector vector which is trainable like other weights of network. All $a_n$ are randomly initialized in the beginning of training. Since multiple $a_n$ can be initialized similarly and it can be inefficient if multiple $A^n$ attend to the same partial states, we explore two regularization methods, entropy and distance, to encourage each $A^n$ attends to different partial states.

$$R_e = \lambda_e * \sum_n \|A^n * \log A^n\| \quad (6)$$

$$R_d = \lambda_d * \exp(-\sum_{n,m}(A^n - A^m)^2) \quad (7)$$

$R_e$ is entropy regularization, which encourages each attention layer to attend to the one partial state. $R_d$ is distance regularization, which encourages each attention layer to attend to different partial states from other attention layers. Those regularization terms are added to loss function, weighted by $\lambda_e$, $\lambda_d$ respectively. We investigate the effects of these regularizations in detail in the experiment section.

**State-action Value Estimation:** Using attention weights from parallel attention layers, weighted value feature is defined as

$$h_n = \sum_i Val_i * A_i^n \quad (8)$$

where $h_n$ is the n-th sum of value features weighted by $A^n$. Then the concatenated feature for state-action value estimation is defined as

$$g = \{h_0, h_1, \ldots, h_N\} \quad (9)$$

The concatenated feature is used to estimate state-action value as follows:

$$Q = f_q(g) \quad (10)$$

where $f_q$ is function approximation. In experiments, we used one fully-connected layer followed by one linear layer as $f_q$. $Q$ has output units for each action so $Q(s,a)$ for all possible actions are calculated with single feed-forward pass.

**Extension to the Multi-Agent Reinforcement Learning**

In multi-agent reinforcement learning tasks, agents perceive partially observed states, which are part of the whole environment state. Agents have to communicate with each other to share information and cooperate to solve the task.

If we use the partially observed states perceived by multiple agents as the partial states of our model, MANet can be extended to the model that can control multiple agents. Using parallel attentions, MANet can learn not only the policy for individual agents but also the communication protocol between agents in an end-to-end fashion. For this purpose, we change modules of our model as follows:

**Input Segmentation:** As mentioned above, we use partially observed states from multiple agents as partial states of our model. Thus, extra segmentation process is not required.

**Feature Extraction:** In the multi-agent setting, each agent should be able to attend to different agents according to their situation. Thus, each agent requires different selector vectors. For this purpose, we extract both key features and selector vectors from inputs of each agent. Equation (2)–(4) are extended as follows:

$$c_i = f_f(s_i) \quad \text{for all } i \in (0,1,\ldots,K) \quad (11)$$

$$Key_i = W_{key} * c_i \quad (12)$$

$$a_i = W_a * c_i \quad (13)$$

$$Val_i = f_v(W_{val} * c_i) \quad (14)$$

where K is the number of agents and $W_a$ is the weight for selector vectors.

**Attentive Communication:** Attention weights for each agent are calculated using

$$A_j^i = \frac{\exp(a^i * Key_j^T)}{\sum_{j'} \exp(a^i * Key_{j'}^T)} \quad i,j \in (0,1,\ldots,K) \quad (15)$$

In this formulation, $A_j^i$ means the attention weight from the agent j to the agent i. Higher $A_j^i$ can be interpreted like the agent j has information which is important for the agent i.

**State-action Value Estimation:** In the multi-agent setting, all agents have their own action so we need state-action values for each agent. For each agent, communication feature and concatenated feature for state-action value estimation are defined respectively as

$$h_i = \sum_j Val_j * A_j^i \quad j \in (0,1,\ldots,K) \quad (16)$$

$$g_i = \{Val_i, h_i\} \quad (17)$$

where $h_i$ is the communication feature for i-th agent. And state-action values for each agent are defined as

$$Q_i = f(g_i) \quad (18)$$

We do not use entropy or distance regularization in the multi agent setting because not like single agent setting, $a_i$ are different for all agents so $A^i$s are unlikely to have similar values. Instead, we add regularization term

$$R = \lambda * (a * Key^T) \quad (19)$$

to the loss function since these values easily diverge without regularization in our experiments.

## Experiments

We designed two game environments, one for single agent task and the other for multi-agent cooperative task. We trained our model as well as the state-of-the-art model and baselines on these environments. Following subsections will cover details about environments and results of experiments.

**Navigation Experiments for Single Agent Task**

**Environment:** For the single agent task, we designed a navigation task in the grid world with waypoints (Figure 1). The map is 8*8 grid and there are 4 waypoints. The agent has to visit the waypoints in an orderly sequence. In each time-step, the agent can move up, down, left, or right. The agent gets reward of 1 when it reaches the proper waypoint, 3 when it has visited all waypoints and -0.01 for each movement. The episode ends when the agent reaches the last waypoint or 200 time-steps have passed. Each cell in the grid is rendered as 5*5 patch so the size of input image is 40*40. The cell that contains the agent is colored green and cells that contain waypoints are colored red. Waypoints that have to be visited earlier have darker color.

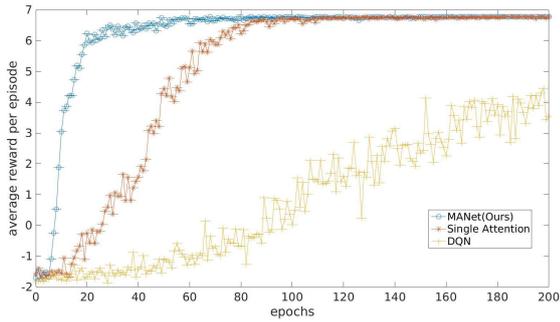

*Figure 3. Learning curves for navigation task.*

**Training Details:** We trained our model, single attention model and DQN. Our model used two attention layers. Details about network structures and hyper parameters are given in supplementary material.

**Quantitative Results:** As shown in the learning curves presented in Figure 3, our model showed great improvement in the training speed and performances. Our model reached near-optimal score (we define near-optimal as score over 6.7) roughly two times faster than the single attention model (Table 1). This result shows that parallel attentions clearly benefits learning complex task. DQN reached near optimal score in only one of three training cases. Furthermore, comparing to the successful case of the DQN, our model reached near-optimal score nearly 3.5 times faster than the DQN (Table 1).

**Qualitative Results and Effect of Regularization:** Examples of the game screenshots and the corresponding attention weights (computed by equation 5) from the proposed model are depicted in Figure 4. As illustrated in Figure 4, in every case of trainings, one of the attention layer mainly attended to the next waypoint and the other one mainly attended to the agent. We also examined attention weights and learning curves of our model with regularizations. As seen in supplementary material, although attention weights of regularized model were more concentrated to one partial state, there was no significant difference in performances. We hypothesize that at least in our domain of experiments, random initialization was enough to make attention layers attend to different partial states. We leave the exploration of other regularization methods and investigation of effects of regularizations in other domains to the future work.

*Table 1. Epochs required for mean score over 6.7*

|  | MANet | Single Attention | DQN |
|---|---|---|---|
| epoch | 57 | 95 | 198 |

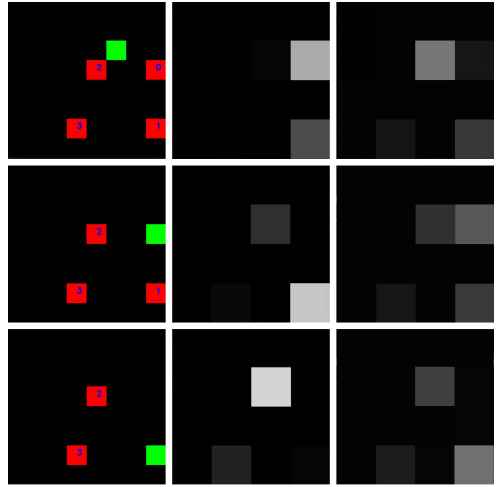

*Figure 4. Game screen(left) and visualization of attentions(middle, right) in navigation task. In all cases of our experiments, one of attention layers attended to the next waypoint (middle) and the other one attended to the agent (right).*

**Combat Experiments for Multi-Agent Task**

**Environment:** To compare our model's performance to baselines and the state-of-the-art model, we implemented modified version of combat environment used in (Sukhbaatar, Szlam, and Fergus 2016). In the combat experiments, two teams battle in 15*15 grid. Each team has five agents. Each team has fixed starting point and agents are spawned randomly in its surrounding 3*3 area when the episode begins. Each agent has 3 health points and can perceive its surrounding 5*5 area. Agents can take 10 actions: moving up, down, left, right, attacking enemy by specifying its index (1~5), or doing nothing. The attack is effective only when the target is in the attacker's surrounding 3*3 area and successful attack reduces target's health point by 1. An agent dies when its health point becomes 0. Agents must have 1 time-step of cool time after the attack. The episode ends when one team wins or 80 time-steps have passed (timeout). The model controls one of the team and the other team is controlled by hard-coded bots. As like (Sukhbaatar, Szlam, and Fergus 2016), our hard-coded bots move randomly until any of the bots spots the enemy. After any of the bots spots the enemy, all bots move to nearest spotted enemy and attack it. The model gets reward of 3.0 when it wins the game and gets reward of -3.0 when it loses or draws (timeout). To encourage agents to attack and accelerate the learning, reward of 1.0 is given for every successful attack and reward of (-1.0 * sum of health points of remaining bots) is given when the model loses or draws. All rewards are shared between agents. Each agent's input is 150-dimensional vector which is concatenation of twenty-five 6-dimensional vectors that encode [x, y, index, team, health, cool-time] of cells in the agent's surrounding 5*5 area. If there is no agent or bot in the cell, the values of index, team, health and cool-time are 0.

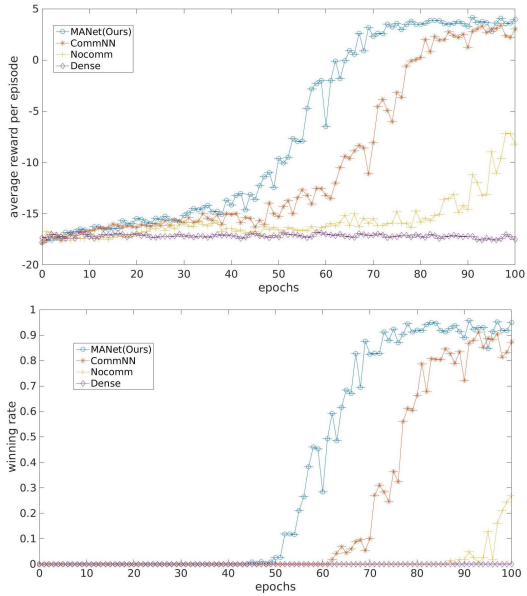

*Figure 5. Learning curves for combat task. The plot on the top is the average reward per episode while the bottom plot is the winning rate.*

**Training Details:** We trained our model, CommNN (Sukhbaatar, Szlam, and Fergus 2016), No-communication model, and Dense communication model (mapping concatenation of inputs of all agents to the state-action values of all agents). Details about the network structures and hyper parameters are given in supplementary material.

**Quantitative Results:** As shown in the learning curves in Figure 5, our model showed roughly 20% faster learning speed than CommNN (Table 2). Though we could not compare the performances accurately because both our model and CommNN reached the near-optimal scores (we define near-optimal as winning rate over 90%), our model showed slightly better performances.

*Table 2. Epochs required for mean winning rate over 90%*

|  | MANet | CommNN | Nocomm | Dense |
|---|---|---|---|---|
| epoch | 73 | 93 | X | X |

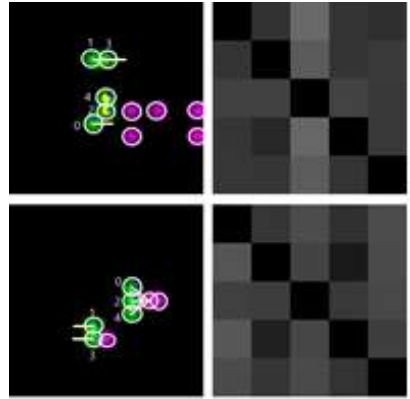

*Figure 6. Game screens (left) and visualization of attention matrixes (right) in the combat task. (i,j)-th cell in attention matrixes is attention weight from the agent j to the agent i. brighter (i,j)-th cell means that the agent i mainly attends to the agent j.*

**Qualitative Results:** Examples of game screenshots and corresponding attention weights from our model are presented in Figure 6. As shown in Figure 6, attention weights were concentrated to the agents which are engaging the enemies (top of Figure 6). This helped our agents to successfully defeat enemies by focusing their fire power on the one enemy at a time (bottom of Figure 6).

## Conclusion

We propose a Multi-focus Attention Network, which can effectively attend to multiple partial states constituting the sensory input to learn faster. Our model shows improved performance and faster learning compared to DQN and the single attention model on the single agent task. Moreover, extended model for multi-agent RL shows faster learning than the state-of-the-art model. In future work, we intend to apply more sophisticated segmentation methods and experiment our model on real world environment. Secondly, we will explore communication between agents having different state-action spaces. Lastly, we will apply parallel attention layers in both spatial and temporal manner. Our final goal is to achieve modular RL model for robot training in the real world.

## Acknowledgments

This research was supported by a grant to Bio-Mimetic Robot Research Center Funded by Defense Acquisition Program Administration, and by Agency for Defense Development (UD130070ID).